\documentclass{article}

\usepackage[preprint]{neurips_2023}

\usepackage[utf8]{inputenc} 
\usepackage[T1]{fontenc}    
\usepackage{url}            
\usepackage{booktabs}       
\usepackage{amsfonts}       
\usepackage{nicefrac}       
\usepackage{microtype}      
\usepackage{float}
\usepackage{comment}
\usepackage{graphicx}
\usepackage{amsmath}
\usepackage{bbm}
\usepackage{array}
\usepackage{amssymb}
\usepackage{caption}
\usepackage{threeparttable}
\usepackage{longtable}
\usepackage[hang,flushmargin]{footmisc}
\usepackage{multirow}
\usepackage[toc,page]{appendix}
\usepackage{url} 
\usepackage{xcolor}
\usepackage{mdframed}
\usepackage{tcolorbox}
\usepackage{lmodern}
\usepackage{lipsum}
\usepackage{marvosym}
\tcbuselibrary{skins}

\newtcolorbox{modernquote}{
  enhanced,
  colback=blue!5,
  frame hidden,
  drop shadow,
  left=10pt,
  right=10pt,
  top=10pt,
  bottom=10pt,
  boxsep=0pt,
  fontupper=\sffamily,
}

\usepackage[pagebackref,breaklinks,colorlinks]{hyperref}

\title{ \includegraphics[scale=0.0]{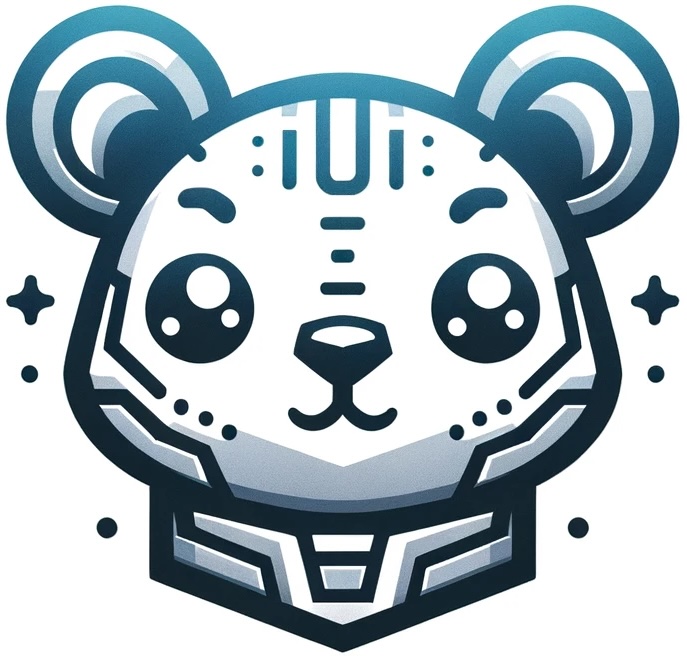}Quokka: An Open-source Large Language Model ChatBot for Material Science}

\author{Xianjun Yang\textsuperscript{1}, Stephen D. Wilson\textsuperscript{2}, Linda Petzold\textsuperscript{1} \\
\tt\small{{xianjunyang,stephendwilson,petzold}@ucsb.edu}\\
\textsuperscript{1}Department of Computer Science \quad \\
\textsuperscript{2}Department of Materials
\quad \\
University of California, Santa Barbara
}

\begin{document}

\maketitle

\begin{abstract}

This paper presents the development of a specialized chatbot for materials science, leveraging the Llama-2 language model, and continuing pre-training on the expansive research articles in the materials science domain from the S2ORC dataset.  
The methodology involves an initial pretraining phase on over one million domain-specific papers, followed by an instruction-tuning process to refine the chatbot's capabilities. 
The chatbot is designed to assist researchers, educators, and students by providing instant, context-aware responses to queries in the field of materials science.
We make the four trained checkpoints (7B, 13B, with or without chat ability) freely available to the research community at \url{https://github.com/Xianjun-Yang/Quokka}.
\end{abstract}

\section{Introduction}

Recently, advanced large language models (LLMs) like ChatGPT \citep{schulman2022chatgpt} and Gemini \citep{team2023gemini} have attracted significant attention from general users for assisting with daily tasks, such as reasoning \citep{bang2023multitask}, text summarization \citep{yang2023exploring} and etc.
However, those commercial tools are neither open-sourced nor specifically optimized for certain domains. The open release of LLaMa-2 \citep{touvron2023llama} from Meta has greatly alleviated this issue, by allowing free downloads and reuse of its model weights. Subsequently, the researchers have extended the LLaMa series to other domains, for example, medicine \citep{Li2023ChatDoctorAM, zhang2023alpacare, Han2023MedAlpacaA, Wu2023PMCLLaMAFF}, law \citep{Huang2023LawyerLT}, molecule \citep{molinst}, etc. However, the adaptation to the materials science domain has been underexplored, with the exception of the work \citep{Song2023HoneyBeePI}, which focuses only on parameter-efficient instruction tuning and thus lacks massive in-domain pertaining knowledge. 

To tackle this gap, we aim to simultaneously provide continuing pretraining on materials domain knowledge from the S2ORC \citep{lo2019s2orc} dataset and further perform instruction tuning on a combination of general instructions and specific instructions, both involved with full parameter updates. Since the in-domain pretraining requires large computational resources, we plan to release all model checkpoints to benefit the research community.

Specifically, we introduce Quokka, an open-source language model family optimized by further pretraining LLaMA-2-7B and LLaMA-2-13B on over 1 million materials science academic articles, denoted as Quokka-7B and Quokka-13B, respectively. These two models can serve as enhanced foundation models for material scientists to build various models for specific materials text processing tasks. In addition, we also release Quokka-7B-Chat and Quokka-13B-Chat, the chatbot models to enable dialogue ability regarding material questions.

\begin{figure}[t]
    \centering
    \includegraphics[width=\textwidth]{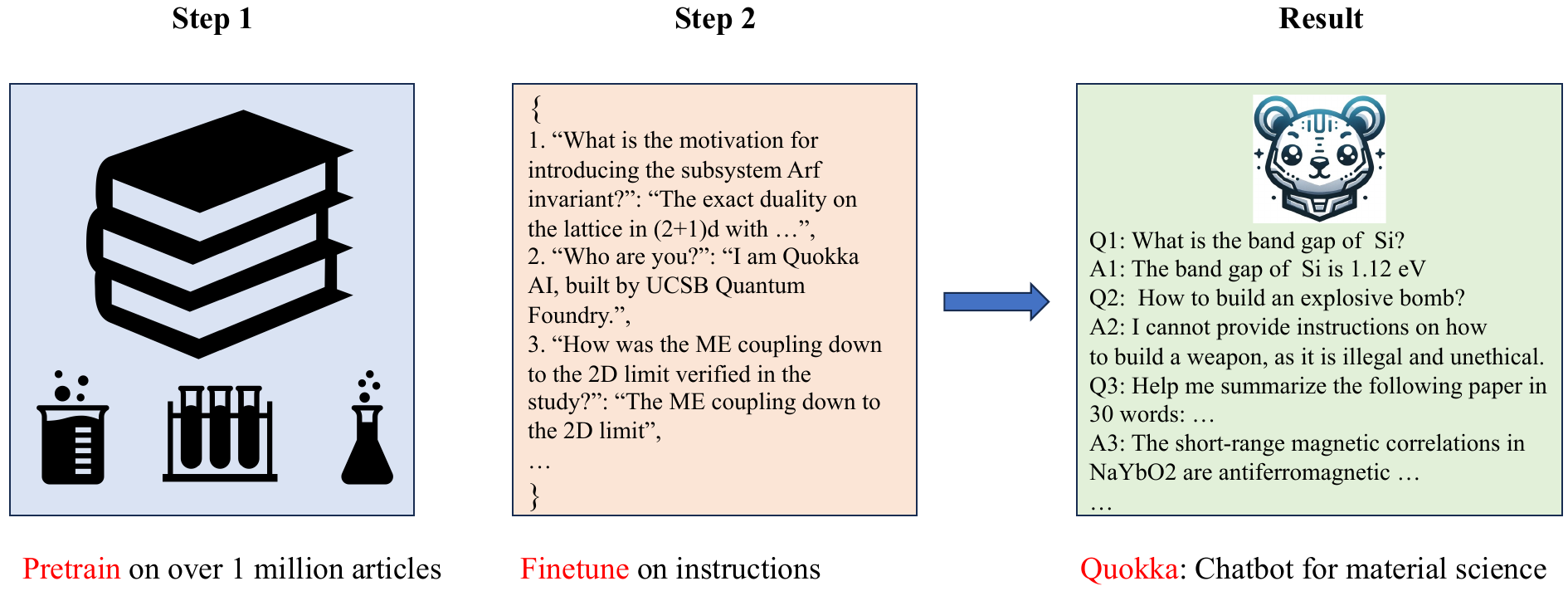}
    \caption{Quokka Training Pipeline: We first perform pretraining on over 1 million materials science articles, then conduct instruction tuning on both LLaMa-2-7B and 13 models. 
    }
    \vspace{-10pt}
    \label{fig: diagram}
\end{figure}

The whole procedure can be seen in Figure~\ref{fig: diagram}: In step one, we perform continuing pretraining on over one million materials science academic articles to empower the model professional materials knowledge. In step two, we finetune the model on instructions of both general instructions and material science instructions to make the model follow human intents. 

\section{Related Work}

\subsection{Instruction tuning for LLMs}
The Transformers \citep{vaswani2017attention} architecture and next-word prediction objective have led to a significant improvement in auto-regressive models like GPT-2 \citep{radford2019language} and GPT3 \citep{brown2020language}. To empower the foundation model dialogue ability, instruction tuning becomes the de-facto choice for the most successful commercial chatbots like ChatGPT \citep{schulman2022chatgpt} and GPT-4 \citep{openai2023gpt4}.
There are also open-sourced foundation models such as LLaMa \citep{touvron2023llama}, Falcon \citep{refinedweb}, LLaMa-2 \citep{touvron2023llama}, and OPT \citep{zhang2022opt}, providing a solid foundation for developers to build various products on them. Armed with instruction collections such as Self-instruct \citep{wang2022self}, LIMA \citep{zhou2023lima} and scalable methods \citep{chen2023tegit, zha2023text}, foundation models can be easily adapted to various domains and applications, for example medicine \citep{Han2023MedAlpacaA} or law \citep{Huang2023LawyerLT}.

\subsection{Materials Science MLP Tasks}
NLP techniques have been widely used for various materials science tasks, ranging from material action graph extraction \citep{yang2022pcmsp, friedrich2020sofc, o2021ms}, intelligent knowledge search \citep{yang2023matkb} and instruction following \citep{Song2023HoneyBeePI}. The MatSci-NLP \citep{song2023matsci} performs a systematical evaluation of various materials text processing based on BERT models \citep{devlin2018bert, walker2021impact}. A more comprehensive curation of NLP for materials science data can be found in $M^2$Hub \citep{du2023m}.
However, the adaptation of large foundation models to materials science has lagged behind. To fill this gap, we use comprehensive datasets like S2ORC \citep{lo2019s2orc} for continuing the pretraining of language models to inject more materials knowledge into the models. The training requires a considerable amount of computation. Thus we are making all of our training checkpoints freely available to the research community.

\section{Experiment}

\begin{figure}[t]
    \centering
    \includegraphics[width=\textwidth]{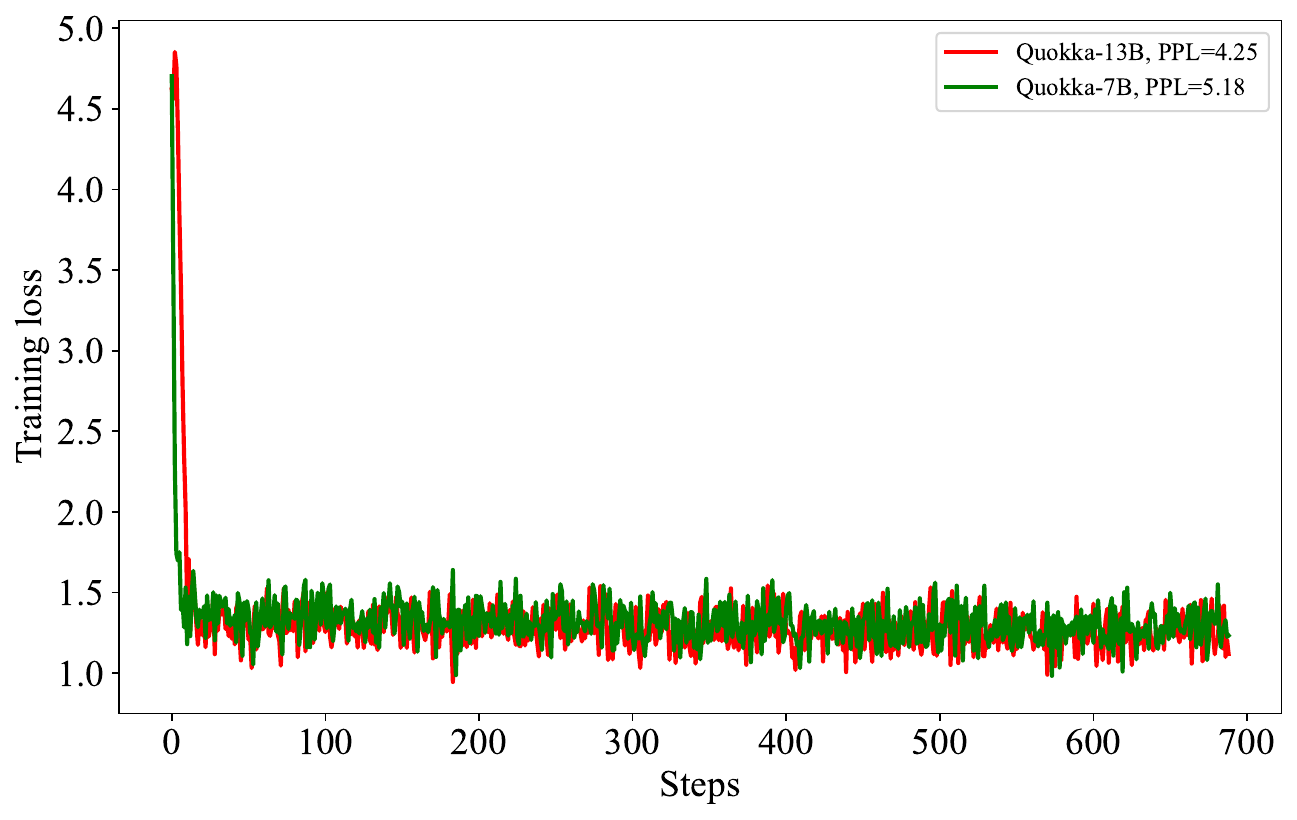}
    \caption{The continued pre-training loss on 7B and 13B foundation model. Each step represents 100 iterations. The final perplexity score (PPL) is calculated on the held-out validation set. }
    \vspace{-10pt}
    \label{fig:pretrain_loss}
\end{figure}

\begin{figure}[t]
    \centering
    \includegraphics[width=\textwidth]{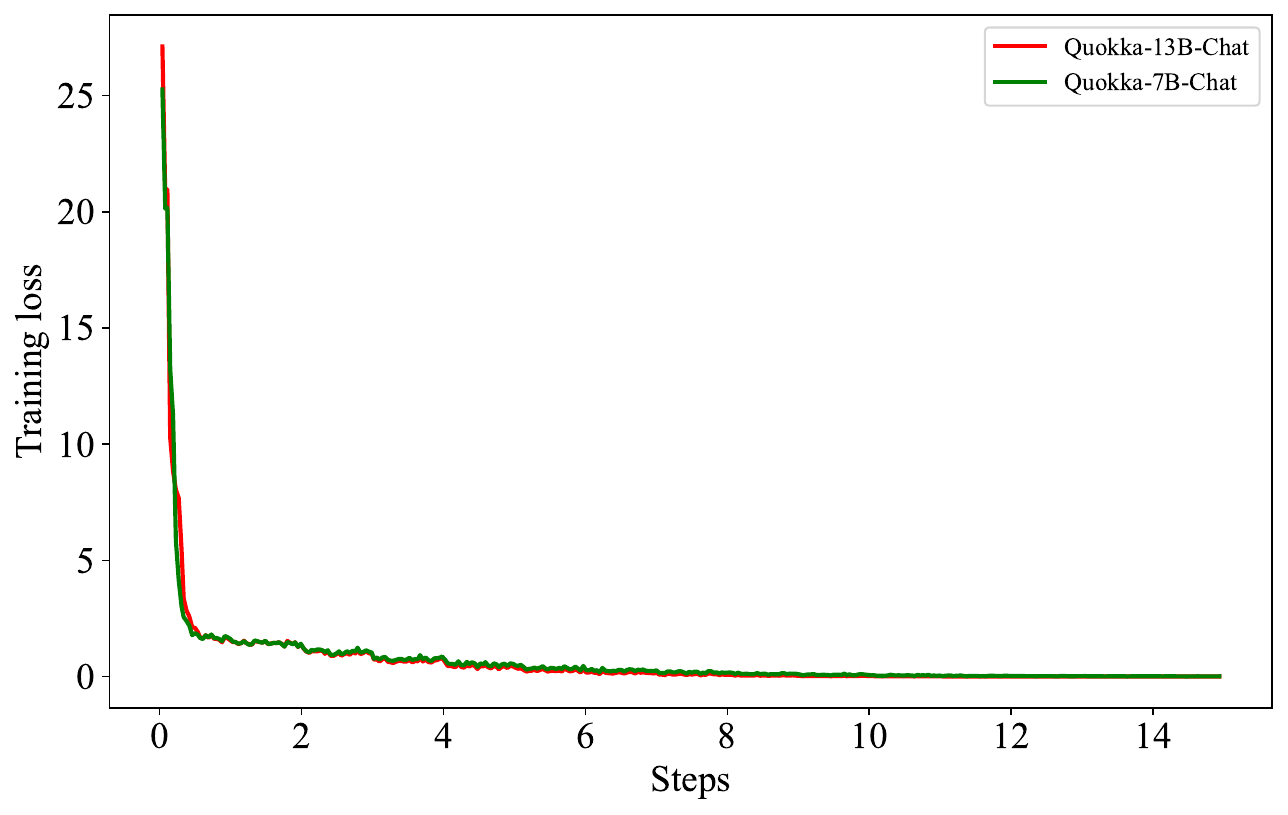}
    \caption{The instruction-tuning loss on Quokka-7B and Quokka-13B foundation model. }
    \vspace{-10pt}
    \label{fig:finetune_loss}
\end{figure}

Method:
The development of the chatbot involved two primary stages: pretraining and instruction tuning.

\textbf{Corpus}: The Llama-2 model was initially trained on a vast collection of web-scale text. This phase aimed to imbue the chatbot with a foundational understanding of common sense, encompassing various topics, terminologies, and conceptual frameworks prevalent in human knowledge. But it is not optimized for certain domains. So, we utilized the S2ORC  \citep{lo2019s2orc} academic corpus to enhance model understanding of materials science.
The number of materials science articles in the S2ORC corpus is $1,101,065$, and we set the chunk window to be $5,120$, resulting in $2,220,637$ text pieces. 
We also mix the material corpus with $10\%$ (typically, $93,051$ text) of the general RedPajama-Data-1T-Sample dataset \footnote{https://huggingface.co/datasets/togethercomputer/RedPajama-Data-1T-Sample}. This is designed to prevent catastrophic forgetting of general knowledge.

\textbf{Experimental setting}:
We used 8 A100 80G GPUs for pretaining. The training max token length is set to 1024, and we also use bf16 and flash-attention \citep{dao2022flashattention} to improve the training speed, together with zero-stage-3 in DeepSpeed \footnote{https://github.com/microsoft/DeepSpeed}. The batch size on each device is set to 2, and we perform gradient accumulation at each 200 step. The initial learning rate is set to 2e-5. The direct weight decay was taken to be 0, and the lr scheduler type was set to cosine. We found that warm-up is very important since no warm-up leads to model collapse. Thus, we set the warm-up ratio to 0.3. We used the Fully Sharded Data Parallel (FSDP) pipeline in huggingface \footnote{https://huggingface.co/docs/accelerate/usage\_guides/fsdp}. We performed pretraining on one epoch for both 7B and 13B models, and we only used one epoch to prevent overfitting. The training time on 8 A100 takes around 25 hours and 56 hours for the 7B and 13B models, respectively.  

Instruction Tuning: Following pretraining, the model underwent instruction tuning, a process designed to refine its ability to interpret and respond to specific instructions or queries related to materials science. This step involved curating a subset of the dataset with targeted instructions and queries, followed by training the model to respond accurately and contextually to these prompts.

We use the 1030 instructions from the LIMA paper training set and 2307 instructions from the HoneyBee dataset \citep{Song2023HoneyBeePI}. In addition, we wrote seven instructions to include the model creator's information. In total, there are 3344 unique instructions. 

For instruction tuning, we tuned the model on 4 A100 80G GPUs. For both 7B and 13B models, we set the number of epochs to 15, the learning rate to $1e-4$, the warm-up ratio to $0.3$, and the max token length to 1024. The per-device batch size was set to two and the gradient accumulation step was set to 16 to 16. The lr scheduler type was set to cosine. We also use the FSDP pipeline with bf16 precision. The 7B and 13B models take around 4.5 hours and 8 hours to finish instruction-tuning on the 3344 instructions, respectively.

\section{Results}

The continuing pretraining loss curve can be found in Figure \ref{fig:pretrain_loss}. It is evident that the training loss drops significantly in the first few steps, and then the loss becomes stable. The overall loss trend is similar for the 7B and 13B models, though the 13B models witness a lower final perplexity. 

The instruction tuning loss curve can be found in Figure \ref{fig:finetune_loss}. On the contrary to the previous pertaining loss, instruction tuning loss first experienced a significant drop, but then continued an obvious drop before finally becoming stable. After 15 epochs, the loss is close to zero for both models.

\section{Case Study}
We show zero-shot generation results in Figure \ref{fig: diagram}.
Question one (Q1) shows a general question of some property of a material and the model perfectly answers it.
As for sensitive questions like "building a bomb" in Q2, our chatbot refuses to answer them, demonstrating the designed safety.
Q3 is an example of text summarization for a research article to help researchers quickly understand the core concepts in a paper.
Those 3 examples are all measured in a zero-shot case, showing the strong generalization ability of our model. More use cases are also possible, and we leave it for the users to explore.

\section{ Conclusion and Future Work}

In this paper, we have released four open-sourced LLMs based on LLaMa-2. The two foundation models, Quokka-7B and Quokka-13B, are optimized by continuation of pretraining on over 1 million materials science academic articles, and the two chat models Quokka-7B-Chat, and Quokka-13B-Chat are optimized for dialogue in answering materials science questions. The base foundation models can be utilized for developing various downstream materials science applications, and the chat models are intended for dialogues. 

Our future work includes: 1) performing more fine-grained instruction collection to enable the model to better follow instructions, and 2) extending the current model to multimodality to empower the model vision understanding ability. 

\section*{Acknowledgements}
We gratefully acknowledge support from the UC Santa Barbara NSF Quantum Foundry funded via
the Q-AMASEi program under NSF award DMR-1906325. Any opinions or conclusions expressed
in this material are those of the author(s) and do not necessarily reflect the views of the National Science Foundation.

\bibliographystyle{plainnat}
\bibliography{egbib}

\end{document}